\documentclass[twocolumn,3p,times,procedia]{elsarticle}

\usepackage{ecrc}
\usepackage{graphicx}
\usepackage{titlesec}
\usepackage{subfigure}
\usepackage{grffile}
\usepackage{color}

\volume{00}

\firstpage{1}

\runauth{}

\jid{procs}

\CopyrightLine{2015}{Published by Elsevier Ltd.}

\usepackage{amssymb}
\usepackage{amsmath}
\usepackage{multicol}
\usepackage[noend]{algpseudocode}

\usepackage{algorithmicx,algorithm}

\usepackage[figuresright]{rotating}
\usepackage{bm}
\usepackage{caption}
\usepackage{float}

\usepackage{fancyhdr}

\fancyheadoffset[RO,EL]{0pt}
\usepackage{geometry}

\begin{document}

\begin{frontmatter}




\dochead{}
\title{
\begin{flushleft}
{\bf \Huge Fast Wireless Sensor Anomaly Detection based on Data Stream in Edge Computing Enabled Smart Greenhouse}
\end{flushleft}
}
 %

\author[]{\bf \Large \leftline {Yihong Yang$^*$$^a$, Sheng Ding$^b$, Yuwen Liu$^a$, Shunmei Meng$^c$, Xiaoxiao Chi$^a$, }
\bf \Large \leftline{Rui Ma$^d$, Chao Yan$^a$}
}

\address{\bf  \leftline {$^a$School of Computer Science,
	 Qufu Normal University, Rizhao 276826, China}

\bf  \leftline {$^b$Shandong Provincial University Laboratory for Protected Horticulture, Weifang University of Science and Technology, }
\bf  \leftline{Shouguang, Weifang, 262700, China}

\bf  \leftline {$^c$Department of Computer Science and Engineering, Nanjing University of Science and Technology, Nanjing 210000, China}

\bf  \leftline {$^d$General Education Department, Shandong First Medical University (Shandong Academy Of Medical Sciences), Taian 271000, China}
}

\cortext[]{Yihong Yang is with the School of Computer Science, Qufu Normal University, China (email: yihongyang8@gmail.com).}

\fntext[]{Sheng Ding is with the Shandong Provincial University Laboratory for Protected Horticulture, Weifang University of Science and Technology, China (email: dingsheng@wfust.edu.cn).}

\fntext[]{Yuwen Liu is with the School of Computer Science, Qufu Normal University, China (email: yuwenliu97@gmail.com).}

\fntext[]{Shunmei Meng is with the Department of Computer Science and Engineering, Nanjing University of Science and Technology, China (email: mengshunmei@njust.edu.cn).}

\fntext[]{Xiaoxiao Chi is with the School of Computer Science, Qufu Normal University, China (email: 925027586@qq.com).}

\fntext[]{Rui Ma is with the General Education Department, Shandong First Medical University (Shandong Academy Of Medical Sciences), China (email: 49803937@qq.com).}

\fntext[]{Chao Yan is with the School of Computer Science, Qufu Normal University, China (email: yanchao@qfnu.edu.cn).}
\begin{abstract}
	
Edge computing enabled smart greenhouse is a representative application of Internet of Things technology, which can monitor the environmental information in real time and employ the information to contribute to intelligent decision-making. In the process, anomaly detection for wireless sensor data plays an important role. However, traditional anomaly detection algorithms originally designed for anomaly detection in static data have not properly considered the inherent characteristics of data stream produced by wireless sensor such as infiniteness, correlations and concept drift, which may pose a considerable challenge on anomaly detection based on data stream, and lead to low detection accuracy and efficiency. First, data stream usually generates quickly which means that it is infinite and enormous, so any traditional off-line anomaly detection algorithm that attempts to store the whole dataset or to scan the dataset multiple times for anomaly detection will run out of memory space. Second, there exist correlations among different data streams, which traditional algorithms hardly consider. Third, the underlying data generation process or data distribution may change over time. Thus, traditional anomaly detection algorithms with no model update will lose their effects. Considering these issues, a novel method (called DLSHiForest) on basis of Locality-Sensitive Hashing and time window technique in this paper is proposed to solve these problems while achieving accurate and efficient detection. Comprehensive experiments are executed using real-world agricultural greenhouse dataset to demonstrate the feasibility of our approach. Experimental results show that our proposal is practicable in addressing challenges of traditional anomaly detection while ensuring accuracy and efficiency.

\end{abstract}


\end{frontmatter}


\section{Introduction}
With the emergence of the information era, Internet of Things (IoT) technology \textcolor{blue}{\cite{li2021lifetime}} has been broadly applied in diverse domains such as intelligent transportation \textcolor{blue}{\cite{dou2020pca}}, smart home \textcolor{blue}{\cite{9372899}} and public security \cite{cai2016collective} \cite{agarwal2021effect}. Edge computing \textcolor{blue}{\cite{9239909}} enabled smart greenhouse is a representative application of Internet of Things technology in the field of agricultural production \textcolor{blue}{\cite{9047920}}, which incorporates edge computing into the smart greenhouse warning system to alleviate the burden of the cloud computing platform \textcolor{blue}{\cite{li2018joint}}. The system mainly consists of cloud computing platform \textcolor{blue}{\cite{xu2020balanced}}, edge computing nodes \textcolor{blue}{\cite{xu2020service}} and wireless sensor nodes \textcolor{blue}{\cite{li2019lyapunov}}, which can monitor the environmental information in real time and employ the information to contribute to intelligent decision-making \textcolor{blue}{\cite{li2015energy}}. As wireless sensor nodes generate data constantly, edge computing nodes first can collect the data and upload it to cloud computing platform \cite{huang2019machine}. The cloud computing platform can allocate tasks to the edge computing nodes. Then, edge computing nodes is to train and upload the local model. Cloud computing platform can fuse local models to obtain the global model and send it to edge nodes. Finally, edge computing nodes utilize the model to detect data in real time \cite{kumari2021analysis}. In the process, detecting the anomalous data collected by edge nodes and reporting it to users to support intelligent decision-making \textcolor{blue}{\cite{xu2020pdm}} are crucial issues. Namely, anomaly detection plays a significant role in the smart greenhouse \cite{xu2020artificial}.

As one of the important research hotspots of data mining \textcolor{blue}{\cite{hu2011emd}}, anomaly detection has been applied to a diverse range of scenarios \cite{zhao2021incremental} such as malware detection, intrusion detection \cite{wang2020research} and event detection in sensor networks \cite{cai2018private} and gained widespread attention from academia and industry \cite{mahmud2020survey} \cite{wang2021robust} \cite{liu2021attention}. Anomaly detection or outlier detection is the recognition of rare items, events or observations. And they can raise our suspicions owing to distinguishing significantly from the majorities or the normal trend. Hawkins \cite{hawkins1980identification} formally defines the notion of an outlier that an outlier is an observation which deviates so much from the other observations as to arouse suspicions that it was generated by a different mechanism. A variety of algorithms have been designed for anomaly detection. Some representative examples are clustering based anomaly detection, distance (or density) based anomaly detection, relative density based anomaly detection, angle based anomaly detection and tree model based anomaly detection.

Wireless sensor data is typical data stream pattern. Data stream or streaming data is increasingly ordinary with the fast advance of Internet of Things and hardware technology \cite{cai2019trading}. It is an infinite sequence of data points with timestamps $t$, which possesses the properties of infiniteness, cross-correlation and concept drift. However, these characteristics pose a considerable challenge on traditional anomaly detection algorithms. Firstly, data stream usually engenders fast which means that it is infinite and enormous. Thus, it is impossible for any traditional anomaly detection algorithm to store the whole dataset or scan the dataset multiple times for the purpose of enhancing accuracy. Secondly, there exist correlations among different data streams, which traditional algorithms hardly consider. In this case, traditional anomaly detection algorithms may perform poorly when executing anomaly detection. Thirdly, the underlying data generation process or data distribution may change over time, which means that the previous training anomaly detection model can’t adapt to the current concepts and predict the subsequent results accurately. Thus, traditional anomaly detection algorithms with no model update will lose their effects.

In consideration of the above challenges, we enhance the current LSHiForest anomaly detection algorithm originally designed for anomaly detection in batched data. Through combining the technique of window, the idea of model update and LSHiForest, a novel anomaly detection algorithm for data stream is proposed, called DLSHiForest, which can be divided into three steps, building initial anomaly detection model, predicting anomaly scores for streaming data points and updating anomaly detection model regularly. Gaining profit from LSH and LSHiForest advantages in efficiency, DLSHiForest can provide accurate and efficient detection.

The main contributions of our paper are threefold:

(1) LSHiForest method is exploited to deal with the problem of excessive time cost in traditional anomaly detection algorithm.

(2) A novel anomaly detection algorithm based on LSHiForest, i.e., DLSHiForest is proposed, which can deal with infiniteness, correlations and concept drift problems.

(3) Extensive experiments are conducted using real-world agricultural greenhouse dataset to confirm the superiority of our approach. Experimental results show that our approach is valid in resolving challenges of traditional anomaly detection while ensuring accuracy and efficiency.

The remaining parts of our paper are designed as below. In Section 2, we elaborate the related work. The motivation of this paper is introduced in Section 3. In Section 4, we first state the LSH and LSHiForest concisely. Then, the proposed DLSHiForest method is formulated thoroughly. For the purpose of convincing the reliability of our method compared with other methods, a large number of experiments are executed in Section 5. Finally, we summarize the paper and reveal our future directions of enrichment.

\section{Related work}

\subsection{Tree model based anomaly detection}
With the development of research in the field of anomaly detection, there exist many approaches associated with tree model \cite{gupta2021prediction}. Isolation forest, a typical tree related outlier detection method, is put forward to detect outliers efficiently. Its core idea is isolating data points to identify anomalies, which relies on the principle that anomalies are sensitive to be isolated compared with normal points. One data point’s anomaly score is the path lengths in the forest \cite{liu2012isolation}. In \cite{tan2011fast}, Tan et al. introduce streaming Half-Space-Trees anomaly detector, specifically for streaming data. During constructing the tree stage, no data need to be provided. And no model reconstruction is demanded so as to adapt to evolving data streams, which makes the method highly efficient. In \cite{wu2014rs}, Wu et al. propose a fast density estimator for streaming anomaly detection, which is carried out by multiple fully randomized space trees (RS-Trees), namely RS-Forest. Random selection of an attribute and a splitting value can construct the RS-Forest. While computing the anomaly score, the node profile that is the number of data points dropping into the node is considered. In \cite{pelossof2009online}, Hoeffding Trees (HT) is proposed to detect streaming data, which requires positive and negative class labels to be obtainable to train model. Nevertheless, the prerequisite is unrealistic because the label of data points is difficult to obtain.

Although numerous tree model based methods have been proposed, there are still a lot of issues to be solved. For example, the algorithm is not capable of dealing with data streams or requires data labels available.

\subsection{\textcolor{blue}{Ensemble learning based anomaly detection}}
Ensemble learning, as one of the research hot topics of machine learning, has been widely applied in the field of anomaly detection, which is combining multiple weak models to obtain an integrated model with strong detection capability. Compared with a single model, the detection and generalization abilities of ensemble learning model perform better \cite{zhang2019delr}. Typical integration methods can be divided into two categories: serial ensemble method and parallel ensemble method. The serial ensemble method is to train multiple classifiers iteratively in series. Each iteration is a process of improving and strengthening the ensemble classifier obtained in the previous iteration. A strong classifier will be obtained at the end of the iteration. The parallel ensemble method is to train multiple weak classifiers independently in parallel, and then combine the results of weak classifiers to obtain a strong classifier \cite{ribeiro2020ensemble}.

In order to solve the problems of existing machine learning based anomaly detection algorithms, Zhong et al. \cite{zhong2020helad} propose a new anomaly detection framework, i.e., HELAD, which is based on the idea of organic integration of various deep learning techniques. The HELAD model combines LSTM classifier and Autoencoder classifier. Experimental results show the superiority of our algorithm. Araya et al. \cite{araya2017ensemble} introduce the ensemble anomaly detection (EAD) framework. The EAD combines several anomaly detection classifiers using majority voting, which include pattern-based and prediction-based anomaly classifiers. Illy et al. \cite{illy2019securing} use the most realistic dataset available for intrusion detection and combine multiple learners to build ensemble learners that increase the accuracy of the detection. The base learner contains random forest classifier and bagging classifier.

Although the above anomaly detection methods based on ensemble learning have their own advantages, they still have some shortcomings such as unexplainable and easy to overfit.

\subsection{\textcolor{blue}{Locality-sensitive hashing based anomaly detection}}
LSH is an algorithm designed for resolving approximate or exact nearest neighbor search in high-dimensional space which is characterized by fast calculation speed \cite{wang2020diversified}. The basic idea is to classify data instances using locality-sensitive functions, for which ‘nearby’ data instances have higher probabilities of being hashed into the identical bucket in comparison with data points that are far away \cite{qi2020privacy}.

Due to its salient features, LSH has recently been employed in anomaly detection field. In \cite{wang2011locality}, a ranking driven outlier detection approach is proposed, which exploits LSH to recognize low-density regions for the sake of refining the data instances in these regions. In \cite{pillutla2011lsh}, Pillutla et al. conduct anomaly detection by utilizing LSH to trim strong inliers, which can reduce the excessiveness of a point in hash tables and make fewer data points used to recognize true outliers. However, the space cost is a problem for the large dataset owing to constructing a number of hash tables. In the above methods, the role of LSH is secondary, just as a pruning technique. On the contrary, Zhang et al. propose a generic framework called LSHiForest combining LSH algorithm and isolation forest model to detect anomaly, in which LSH plays an important role instead of parameter tuning \cite{zhang2017lshiforest}. In addition, the instantiation of the framework can be achieved with various LSH such as angle-based LSH, kernelized LSH and  $l_p$  ($p$ = 1, 2)  LSH. This work has been proven to have better performance in terms of accuracy, efficiency and robustness. Thus, our work is carried out based on LSHiForest.

\section{Motivation}
The example is exploited to explain our proposal’s motivation in Fig.1. Edge computing enabled smart greenhouse is a typical application of Internet of Things technology in the field of agricultural production. The edge computing enabled smart greenhouse refers to the real-time monitoring of temperature, humidity, light, carbon dioxide concentration and other information in the greenhouse through edge computing nodes and employs these information to contribute to intelligent decision-making supported by warning system. In this way, anomaly information is reported to the user. Furthermore, rolling shutters, lights, irrigation and other equipment in the greenhouse could be controlled automatically and the crops in the greenhouse could grow in a suitable environment. As a result, the yield and quality of crops could be enhanced and the revenue also increases. In this process, detecting the data collected by edge nodes, discovering the abnormal data, and reporting it to the user to support intelligent decision-making are crucial issues. Therefore, this motivates our research. However, there are often two challenges arising from this as follows:

(1) The data in the system is continuously generated over time and shows the characteristics of infiniteness, correlations and concept drift, which makes traditional anomaly detection algorithms for static data ineffective.

(2) The anomaly detection algorithm in data stream is confronted with the dilemma of accuracy and efficiency as well.

In view of the above problems, the anomaly detection method, named DLSHiForest, is proposed to achieve accurate and efficient detection goals while taking into account the features of data stream.

\begin{figure}
	\centering
	\includegraphics[width=1.0\linewidth]{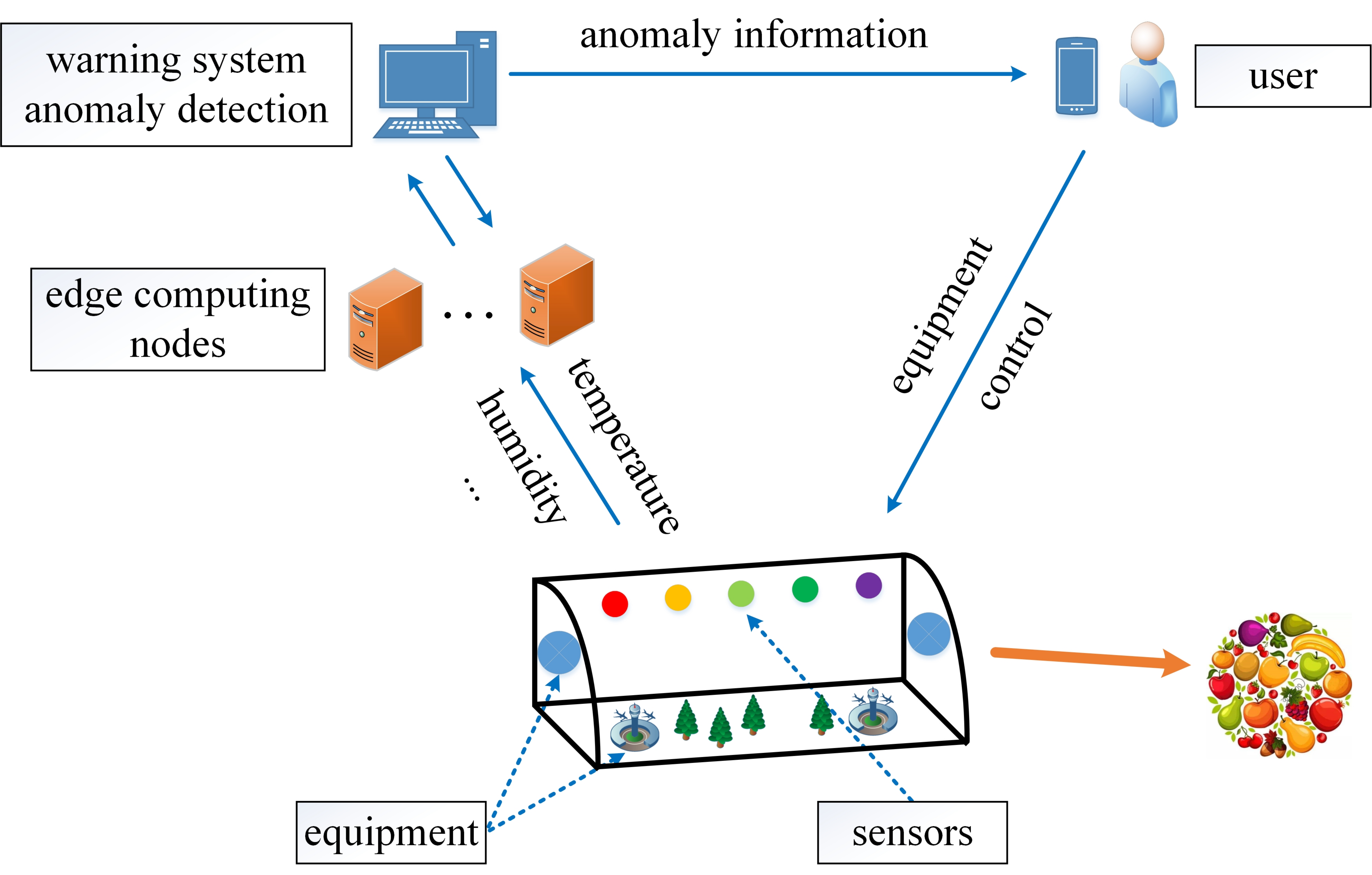}
	\caption*{Fig. 1: An example of anomaly detection in smart greenhouse}
\end{figure}

\section{The detection method}
We present LSHiForest briefly in section 4.1. In section 4.2, our approach DLSHiForest is elaborated in detail.
\subsection{LSHiForest}
Locality Sensitive Hashing (LSH) is an efficient similarity search method, of which the basic operation is to hash data points into the buckets using locality sensitive hashing functions \cite{wang2020diversified1}. Different buckets are distinguished by the hash values after hashing data points \cite{khazbak2020preserving}. The rationale behind the operation is that close or similar data points have higher probability of being hashed into the same bucket than points that are far or dissimilar. Given a distance metric, $d_1<d_2$ are two distance values, and $p_1, p_2, 0 \le p_1, p_2 \le 1$ are two probability values. A hash function family $\digamma$ is said to be $(d_1, d_2, p_1, p_2)$-sensitive if for any function $f \in \digamma, x, y \in D$, the following conditions hold \cite{qi2020privacy2}:
\begin{equation}
d\left( {x,y} \right) \le d_1 \Rightarrow Pr\left[ {f(x) = f(y)} \right] \ge p_1
\end{equation}
\begin{equation}
d(x,y) \ge d_2 \Rightarrow Pr\left[ {f(x) = f(y)} \right] \le p_2
\end{equation}

The equation $f(x)=f(y)$ refers that $x$ and $y$ are hashed into the identical bucket of a hash table. The condition $p_1>p_2$ is usually demanded to make an LSH family useful.

LSH is usually applied in the domain of anomaly detection on account of its above salient features. LSHiForest is one of the typical applications of LSH, which incorporates LSH into isolation forest. The core idea of LSHiForest is to find outliers by partitioning. Like in iForest, the LSHiForest also consists of training stage and testing stage. In the training stage, the main task is to construct LSHiForest. Given an LSH family, we build an LSHiTree by means of recursively hashing a subsample $S$$\in$$D$. The recursive process stops when all data points are separated or division reaches a height limit. And numerous such trees make up the LSHiForest. In addition, the sample is obtained by sampling data points in the dataset. The sampling rate is $min\left\{1,\frac{2^s}{n}\right\}$, where $s$ follows the uniform distribution $U\left(6,10\right)$. Specifically, $\psi$ will change in [64,1024] if $n$$\ge$64. And the tree height limit can be estimated by the average height of a digital trie since an LSHiTree can be regarded as a digital trie. There exist random digital tries with $n$ data instances, over a digit alphabet $\left\{K_1,\cdots,K_v\right\}$, $v\ge2$. To constitute a sequence of digits for data point indexing, $K_i$ is chosen for an element of the sequence with probability $p_i$, $1 \le i \le v$. The average height of the digital tries avg($n$) is about $[2ln(n)+\gamma-ln(2)]/g+1$, where $\gamma  \approx 0.5772$ is the Euler constant and ${\rm{g}} =  - \ln (\sum\nolimits_{i = 1}^v {p_i^2} )$. Assuming all data points in $D$ are independent, and the hash values which an LSH family engenders are also independent and of equal probability, i.e., $p_i = \frac{1}{v}$. Then, a proper upper limit of $H$ can be:

\begin{equation}
	E(H) \approx avg(\psi) \le 2\log 2(\psi ) + 0.8327
\end{equation}

In the testing stage, the main task is computing outlier scores for data instances. For a data instance $x$$\in$$D$, the sum of the path lengths that are obtained by traversing each tree in the forest is treated as final anomaly score. Before the combination of path lengths, $\mu (\psi )$ is used to normalize $h_i\left(x\right)$.
\begin{equation}
\mu (\psi ) = \left\{ {\begin{array}{*{20}{c}}
			{\frac{{\ln (\psi ) + \ln (v - 1) + \gamma }}{{\ln (v)}} - \frac{1}{2},}&{\psi  > v}\\
			{1,}&{1 < \psi  \le v}\\
			{0,}&{\psi  \le 1}
	\end{array}} \right.
\end{equation}

Moreover, the exponential function $2^{-x} (x\ge0)$ is exploited to scale the normalized path lengths into the interval (0, 1]. Thus, the ultimate anomaly score of data point $x$ is:
\begin{equation}
	AS_x = \frac{1}{t}\sum\limits_{i = 1}^t {{2^{ - \frac{{h_i(x)}}{{\mu(\psi_i)}}}}} 
\end{equation}

We explain the essential idea of LSHiForest in Fig.2. Suppose that $a, b, c, d \in D$ represent a sample. Before building a tree, we firstly generate a set of hash functions $hash_0$, $hash_1$, $\cdots$ , $hash_{50}$, i.e., hash family. Then, we create the first node, namely, the root node, and put $a, b, c, d$ into the node. Subsequently, $hash_0$ is employed to hash data points $a, b, c, d$ and get different hash values. According to the different hash values, $a, b, c, d$ are divided into different subsets that represent different child nodes. Similarly, $hash_1$ is used to partition the previously generated subsets. And we repeat the process till all data instances are separated or partition reaches a height limit, which is the building steps of a tree. If we construct multiple trees, the LSHiForest will be obtained. In this way, the training process of the tree model is completed. In the testing stage, a data point traverses each tree in the forest through LSH functions while the path length in the tree is recorded. After the traversal is completed, we get the final anomaly score of the data point.

\begin{figure}
	\centering
	\includegraphics[width=1.0\linewidth]{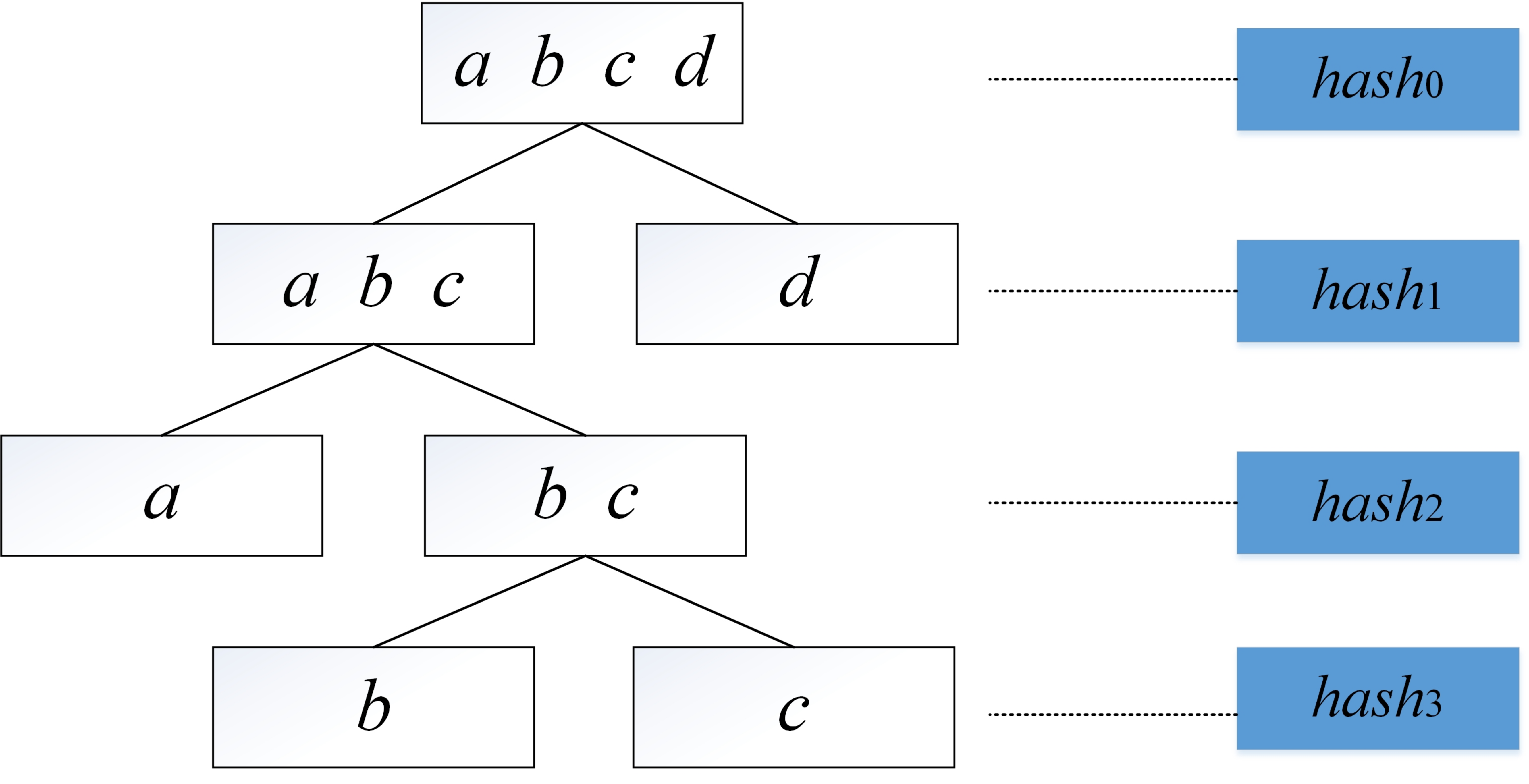}
	\caption*{Fig. 2: The process of building LSHiForest}
\end{figure}

It can be seen from the aforementioned procedure that the LSHiForest can achieve efficient and accurate anomaly detection. Thus, we apply LSHiForest to anomaly detection about data stream to detect anomalies.

\subsection{DLSHiForest: Dynamic anomaly detection based on LSHiForest}
In this section, a new anomaly detection method DLSHiForest is proposed. The key principle is: on basis of LSH technique and the first $w$ data points in the data stream, we build multiple trees as the initial anomaly detection model; afterward, we compute the anomaly score for each streaming data point by traversing each tree; finally, we update the model regularly utilizing the current data points in the window. Specifically, as is presented in Fig.3, our method comprises three stages. In Tab.1, symbols and their meanings in our paper are displayed.

\begin{table}[!htbp]	
	\begin{center}
		\noindent\fbox{\parbox{.9\linewidth}{
				\textbf{Step 1: Building initial anomaly detection model.}\\ 
				\hspace{1em} Ground on the first $w$ data points in the data stream and generated multiple sets of hash functions, build initial tree model.\\
				\textbf{Step 2: Predicting anomaly scores for streaming data points.}\\
				\hspace{1em} For each streaming data point, get the path length through traversing the tree in the model. Then, we get the final anomaly score by computing the sum of the
				path lengths.\\
				\textbf{Step 3: Updating anomaly detection model regularly.}\\
				\hspace{1em} While the window is filled, we utilize data points in
				the window to update the model to adapt to the concept drift.}
		}
		\caption*{Fig. 3: Three steps of the DLSHiForest method}
	\end{center}
\end{table}

\textbf{Step 1: Building initial anomaly detection model.}

In this step, we utilize the first $w$ data points in the data stream and hash functions generated in advance to build multiple LSHiTrees as the initial anomaly detection model. Here, $x_i$ denotes a certain data point, \textcolor{blue}{which we regard as a multidimensional streaming data point, i.e., each dimension of the data point is a single data stream. Thus, our approach works on several data streams contemporarily}

\begin{table}	
	\centering
	\caption*{Tab. 1: Symbols and meanings}
	\begin{tabular}{c|l}
		\hline
		$x$ & a data point\\
		\hline
		$D$ & dataset\\
		\hline
		$n$ & number of data points in the dataset\\
		\hline
		$m$ & number of dimensions\\
		\hline
		$t$ & number of trees\\
		\hline
		$w$ & window size\\
		\hline
		$b$ & subset size\\
		\hline
		$S$ & a subsample of $D$\\
		\hline
		$\psi$ & sample size\\
		\hline
		$v$ & branching factor\\
		\hline
		${\rm{f}}( \cdot )$ & hash function\\
		\hline
		$\digamma$ & hash function family\\
		\hline
		$h( \cdot )$ & path length\\
		\hline
		$H$ & tree height limit\\
		\hline		
	\end{tabular}
	
\end{table}

First, we select the first $w$ data points in the data stream as the sample to build trees. For each tree, a set of hash functions, namely, a hash family is produced ahead of constructing the tree. \textcolor{blue}{Then, for each streaming data point $x_i$ in the sample, we regard it as an $m$-dimensional vector $\overrightarrow {x_i}  = (x_{i,1},x_{i,2}, \cdot  \cdot  \cdot ,x_{i,m})$. Each dimension of the streaming data point is a single data stream. Multiple data streams form a data point. The dimension of data streams refers to the dimension of a streaming data point.} The tree construction is a recursive division process. At every procedure of segmentation, a hash function $f_I$ of LSH family is utilized to engender hash values for all data points. \textcolor{blue}{We use different hash functions for different node to find out outliers as soon as possible. Because the partition results are various for different hash functions. If we always use the same hash function, the partition result obtained is always identical, which is impossible to isolate outliers.} For a certain data point, the hash function considers all the dimension information while hashing it, namely, our method takes into account cross-correlation among different streams. \textcolor{blue}{The correlation of data stream refers to mutual influence between different attributes of a data point. For example, a data point contains temperature and humidity attributes. The value of humidity will decrease as the temperature value increases. There exists relevant relationship between them, and they have an effect on each other, which is the correlation between data streams. The dimension of hash function is equal to the dimension of data point, which can be regarded as two vectors. The process of hashing is the dot produce of these two vectors, which indicates that the hash function operates on all dimensions of the data point meanwhile rather than operation only on a single dimension of the data point. Namely, the hash function takes into account all dimension information of the data point. And the hash function considers correlation.} Finally, the data points are divided into non-overlapping subsets related to hash values. Besides, to produce a compact tree without single-branch paths, the data is split repeatedly until generating multiple hash values or reaching the tree height limit.  

\begin{algorithm}
	\caption{BuildLSHiTree ($w, H, I, X, \digamma$)} 
	{\bf Input:		
	} 
	$w$ - window size\\
	\hspace*{1.15cm}$H$ - height limit\\
	\hspace*{1.15cm}$I$ - index\\
	\hspace*{1.15cm}$X$ - input data\\
	\hspace*{1.15cm}$\digamma$ - LSH family\\
	{\bf Output:} 
	$T$ - an LSHiTree
	\begin{algorithmic}[1]
		\If{$w$ == 0} 
		\State \Return NULL
		\EndIf
		\If{$w$ == 1 OR $I > H$ } 
		\State \mbox{\Return node$\{$size = $w$, hash\_index = $I$, children = $\phi$$\}$}
		\Else
		\State $\{ K_1:X_1, …, K_v:X_v \}$ = lsh\_split($X, f_I$), $f_I \in \digamma$
		\While{$v$ = 1 AND $I \leq H$} 
		\State{$I = I + 1$
			\State$\{K_1:X_1, …, K_v:X_v\}$ = lsh\_split($X, f_I$)} 
		\EndWhile
		\If{$I \ge H$} 
		\State \Return node$\{$size=$w$, hash\_index=$I$$\}$
		\EndIf	
		\State initialize child node set: $C =\phi$
		\For{$i, 1 \leq i \leq v$} 
		\State $c_i$ = BuildLSHiTree($w, H, I+1, X_i, \digamma$)
		\State $C = C\cup c_i$
		\EndFor
		\State \mbox{\Return node$\{$size = $w$, hash\_index = $I$, children = $C\}$}
		\EndIf
	\end{algorithmic}
\end{algorithm}

The above is the process of building a tree, which is represented in Algorithm 1. Moreover, the anomaly detection model is comprised of multiple trees, which requires repeating the Algorithm 1 many times.

\textbf{Step 2: Predicting anomaly scores for streaming data points.}

In Step 1, the initial anomaly detection model has been established, which can be exploited to detect outliers for the continuously arriving streaming data points. We elaborate the process of detection in this step.

While the data point arrives, it traverses the tree in the model to get the path length. First, the hash function in the root node is to hash data point to acquire the hash value that determines the data point is put into which child nodes. Then, the hash function corresponding to the child node continues to hash the data point, which is repeated until the data point arrives the corresponding node. During the traversal, the current depth and corresponding hash function index of the data point are recorded to compute the path length, which is represented in Algorithm 2. The current depth and hash function index record the compressed and uncompressed path length information of a data point respectively. In addition, due to the infiniteness of data stream, the whole dataset can’t be stored as the continuous arrival of data points. Thus, we set the window that is fixed size to store the latest data. \textcolor{blue}{The window refers to the sliding window, which consists of two sliding endpoints. And it can store a certain amount of real-time data in time, the end of which is always the current timestamp. While the sliding window continuously receives newly generated data, it will delete old data.}

\begin{algorithm}
	\caption{PathLength ($x, node, d, \digamma, g$)} 
	{\bf Input:		
	} 
	$x$ - data point\\
	\hspace*{1.15cm}$node$ - current node\\
	\hspace*{1.15cm}$d$ - current depth\\
	\hspace*{1.15cm}$\digamma$ - LSH family\\
	\hspace*{1.15cm}$g$ - granularity adjustment factor\\
	{\bf Output:} 
	The path length of $x$ in the tree $T$
	\begin{algorithmic}[1]
		\If{$node$ == NULL} 
		\State \Return -1
		\ElsIf{$node$.children == $\phi$}
		\State e = ($node$.hash\_index)/$d$
		\State $\rm{temp_2}$ = $d \ast \rm{e}^g$
		\State \Return $\rm{temp_2}$ + $\mu$($node$.size)
		\Else
		\State $K$ = $f_{hash\_index}$($x$), $f_{hash\_index} \in \digamma$
		\If{$\exists(K_i:X_i)\in node$.children AND $K=K_i$}
		\State \Return PathLength($x, node, d+1, \digamma, 1$)
		\Else
		\State $node$.hash\_index = $node$.hash\_index + 1 
		\State  $\rm{e}$ = ($node$.hash\_index)/($d$+1)
		\State \Return ($d$+1) $\ast \rm{e}^g$	
		\EndIf
		\EndIf
	\end{algorithmic}
\end{algorithm}

The above is the path length of single tree. And the final anomaly score for the data point needs to repeat the Algorithm 2 many times to get the sum of path lengths in all trees of forest. \textcolor{blue}{We regard data points the anomaly score of which is greater than the threshold as anomalies. Since the value of threshold is different in various fields, we set it to 0.65 in our paper through a set of experiments and relevant expert experience.} 

\textbf{Step 3: Updating anomaly detection model regularly.}

Because of the existence of concept drift in the data stream, anomaly detection model needs to be updated regularly to adapt to the changes over data distribution or data generation process. We employ the data points in the window to amend the model.  

First, with continuous arrival of data points, the window of size $w$ stores the data points in time. Then, while the number of data points stored is equal to $w$, the data points in the window are utilized to update the detection model, which can be achieved by calling the Algorithm 1 many times. Last, while the model update is accomplished, the window would be cleared and the next point could be handled immediately. \textcolor{blue}{Each window does not overlap with previous data points. Because data enters into the window in the form of data block rather than a single data point. Data block consists of multiple data points and the sliding window can contain a data block in our paper. The sliding of the window indicates the inflow of new data block and the deletion of old data block.}

The above is the procedure of model update. And Algorithm 3 describes the entire process of our method in detail.

\begin{algorithm}
	\caption{DLSHiForest ($t, w, X$)} 
	{\bf Input:		
	} 
	$t$ - number of trees\\
	\hspace*{1.15cm}$w$ - window size\\
	\hspace*{1.15cm}$X$ - input data\\
	{\bf Output:} 
    \mbox{$AS_x$ - anomaly score for each data point $x$}
	\begin{algorithmic}[1]
		\For{$i$, $1 \le i \le t$}
		\State compute a tree height limit $H_i$
		\State BuildLSHiTree($w, H_i, 0, X, \digamma$) 
		\EndFor
		\State Count = 0
		\While{data stream continues}
		\State Receive the data point $x$
		\State $AS_x$ = 0
		\For{each tree $T$ in Forest}
		\State temp=PathLength($x, node, d, \digamma, 1$)/ $\mu$($w$)
		\State  $AS_x=AS_x+2^{-temp}$   
		\EndFor
		\State Report $AS_x$ as the anomaly score for $x$
		\State Count ++
		\If{Count == $w$ }
		\For{$i$, $1 \le i \le t$}
		\State compute a height limit $H_i$
		\State BuildLSHiTree($w, H_i, 0, X, \digamma$)
		\EndFor
		\State Count = 0
		\EndIf
		\EndWhile
	\end{algorithmic}
\end{algorithm}

\section{Experiments}

\subsection{Experimental settings}
We implement extensive experiments based on real-world agricultural greenhouse dataset \textcolor{blue}{(https://github.com/yangyihong/dataset)} in the part. The dataset contains six attributes, namely indoor temperature, indoor humidity, indoor lighting, indoor carbon dioxide, soil temperature and soil humidity. To eliminate the dimensional influence between attributes, all attributes are normalized using RobustScaler. While running each experiment, we randomly select a subset as the input of algorithm for the sake of running time. And the subset size is denoted by $b$. \textcolor{blue}{The subset refers to a continuous piece of data in the dataset, which we obtain by randomly selecting data points of preset size from dataset. The selected subset is different in each time.} Moreover, it has been demonstrated that LSHiForest shows best comprehensive property with regard to accuracy and efficiency when using $l_p$ ($p$ = 2) LSH. Thus, we utilize $l_p$ ($p$ = 2) LSH to hash data in our experiment. All experiments are executed on a Lenovo PC with 2.6 GHz processors, 8.0 GB RAM and Windows 10 (64bits) operating system. And we deploy all the code in Python 3.6. We run each experiment 60 times, then treat the average experimental outcome as final result.
\subsection{Evaluation metrics}
This paper chooses three metrics to evaluate the detection performances:

(1) AUC (Area Under Curve): we use AUC to measure the accuracy of algorithms. AUC is defined as the area between ROC curve and coordinate axes. The larger the AUC is, the higher the detection accuracy would be.

(2) F1-Score: F1-Score is also utilized to evaluate our algorithms’ accuracy. As shown in Eq.(6), it is an estimator to estimate accuracy of the two-class model in the statistics, which takes account of precision and recall. As F1-Score gets larger, our detection accuracy would become higher.
\textcolor{blue}{
\begin{equation}
	F1-Score = 2 \cdot \frac{{precision \cdot recall}}{{precision + recall}}
\end{equation}
}

(3) Time Cost: time consumed is employed to assess the detection efficiency. The timer begin to time as soon as an incoming data point arrives, and stop timing while the anomaly score of the data point is obtained. Our detection efficiency would get higher as Time Cost becomes lower.

\subsection{Comparison methods}
We evaluate the detection performances of proposed anomaly detection algorithm by comparing it with the following three methods:

(1)	\textbf{Hyper\_grid} \cite{ding2017streaming}: the hyper-grid structure and online ensemble learning technique is exploited to detect anomalies for streaming data.

(2)	\textbf{Cluster} \cite{yin2019anomaly}: it improves the existing data stream clustering algorithm and designs the anomaly detection model on the basis of improved algorithm to detect outliers of data stream.

(3)	\textbf{RRCF} \cite{guha2016robust}: it exploits robust random cut forest data structure to conduct anomaly detection in a dynamic data stream. And the data structure can be viewed as a sketch or synopsis of the input stream.

\begin{figure}[!t]
	\centering
	\subfigure[AUC]{
		\includegraphics[width=1.0\linewidth]{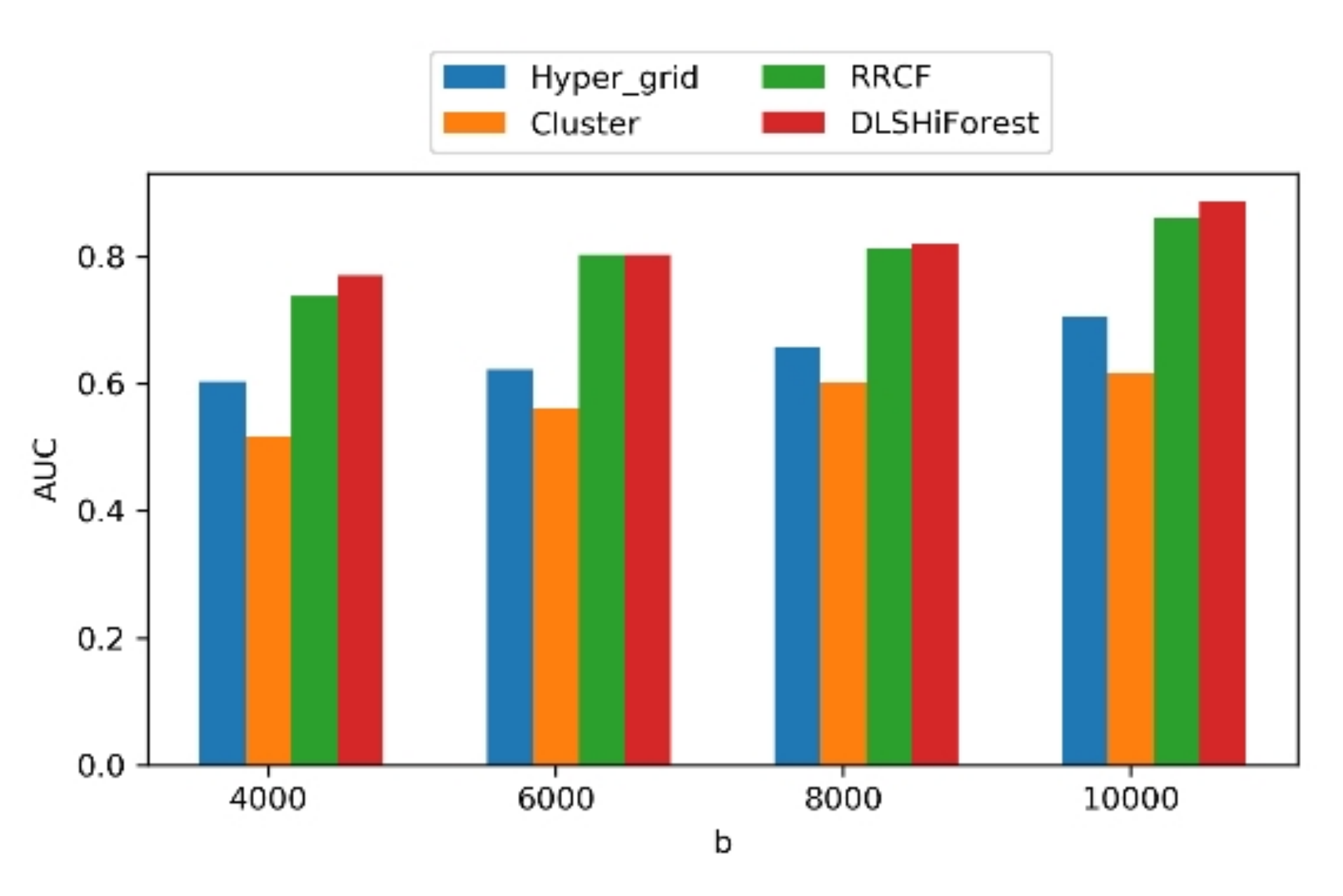}
	}
	\quad
	\subfigure[F1-Score]{
		\includegraphics[width=1.0\linewidth]{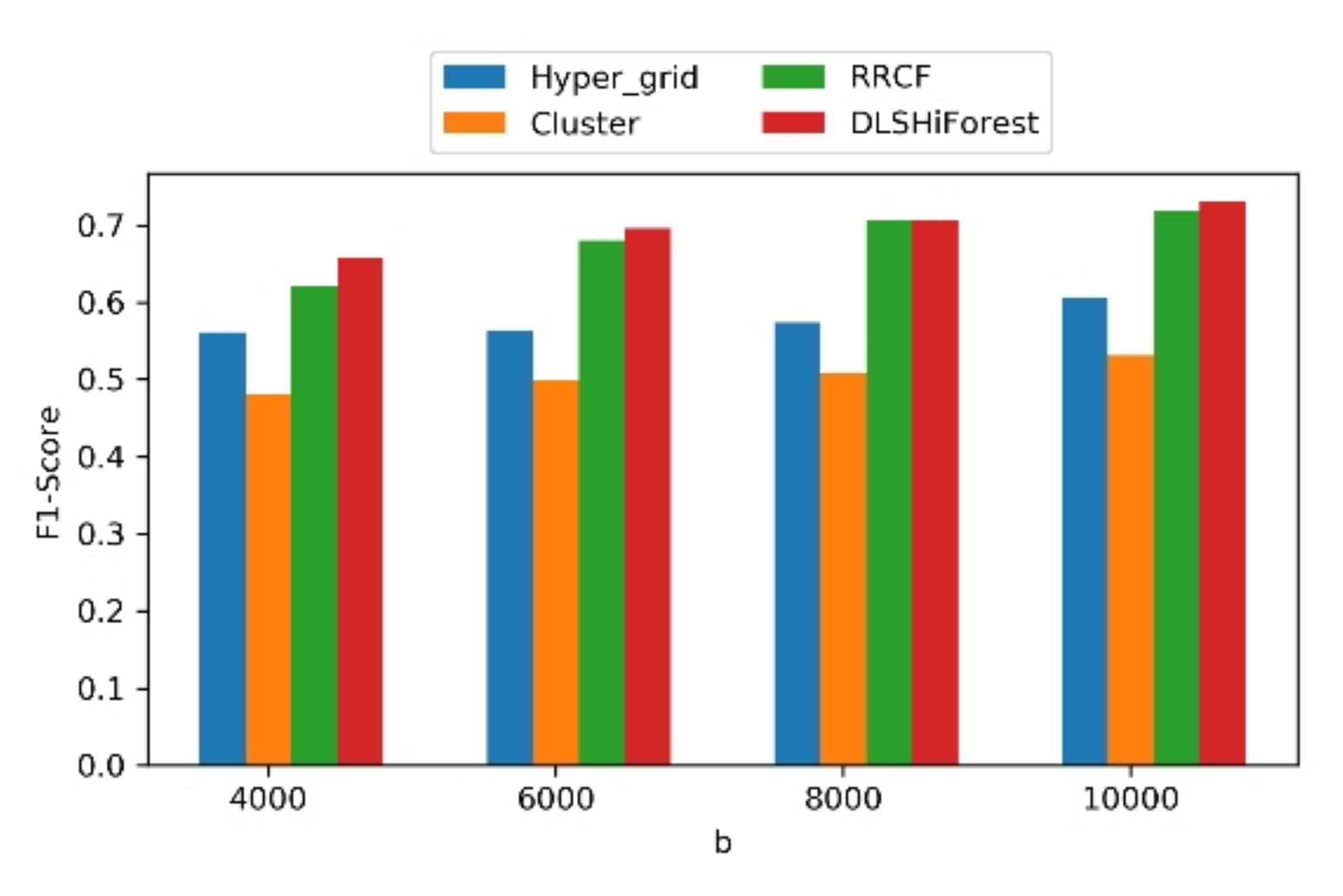}
	}	
	\caption*{Fig. 4: Detection accuracy comparison}
\end{figure}

\subsection{Experimental results}
In order to gauge our approach, five profiles are scheduled and tested in this section. Here, $w$ represents the window size, $t$ represents the number of trees as depicted in Tab.1.

\textbf{Profile 1: Accuracy comparison of the four methods}

The AUC and F1-Score metrics are evaluated to assess the accuracy of four methods in this part. Here, $w$ = 128, $t$ = 60, $b$ = $\left\{4000, 6000, 8000, 10000\right\}$. The experimental results are shown in Fig.4.

As exhibited in Fig.4, our approach maintains a stable and competitive detection performance in terms of both AUC and F1-Score. LSHiForest is a tree isolation based ensemble anomaly detection method combined with LSH, which can apply to various situations. Thus, the most outliers could be found based on the LSHiForest, which is more beneficial to detection quality. The performance of RRCF is slight inferior as it partition data points by randomly selecting dimension and division point, which exists a certain degree of randomness. Besides, Hyper\_grid and Cluster are less accurate than RRCF. This is due to the fact that they hardly consider the correlation between the different dimensions.  

The AUC and F1-Score acquired from accuracy detection of four approaches all display a growing tendency as the growth of value $b$. This reason is that the more information is accessible as the subset size rises.

\textbf{Profile 2: Efficiency comparison of the four methods}

In order to compare four methods’ efficiency, we measure the Time Cost metric in this part. Here, $w$ = 128, $t$ = 60, $b$ = $\left\{4000, 6000, 8000, 10000\right\}$. The experimental outcomes are shown in Fig.5.

As revealed in Fig.5, our approach performs primely in terms of efficiency. The reason why our method is more efficient is the use of LSH when hashing points for traversing the tree. Thus, the anomaly score could be obtained in time. RRCF also shows better performance compared with Hyper\_grid and Cluster due to the salient features of tree structure. However, Hyper\_grid and Cluster perform worse which may result from the inherent features of grid and cluster structure.

The four methods’ Time Cost remains relatively unchanged with the increase of $b$. Because the calculated time refers to the time spent in the testing stage, which is less affected by the subset size.

\begin{figure}[!htbp]
	\centering
	\includegraphics[width=1.0\linewidth]{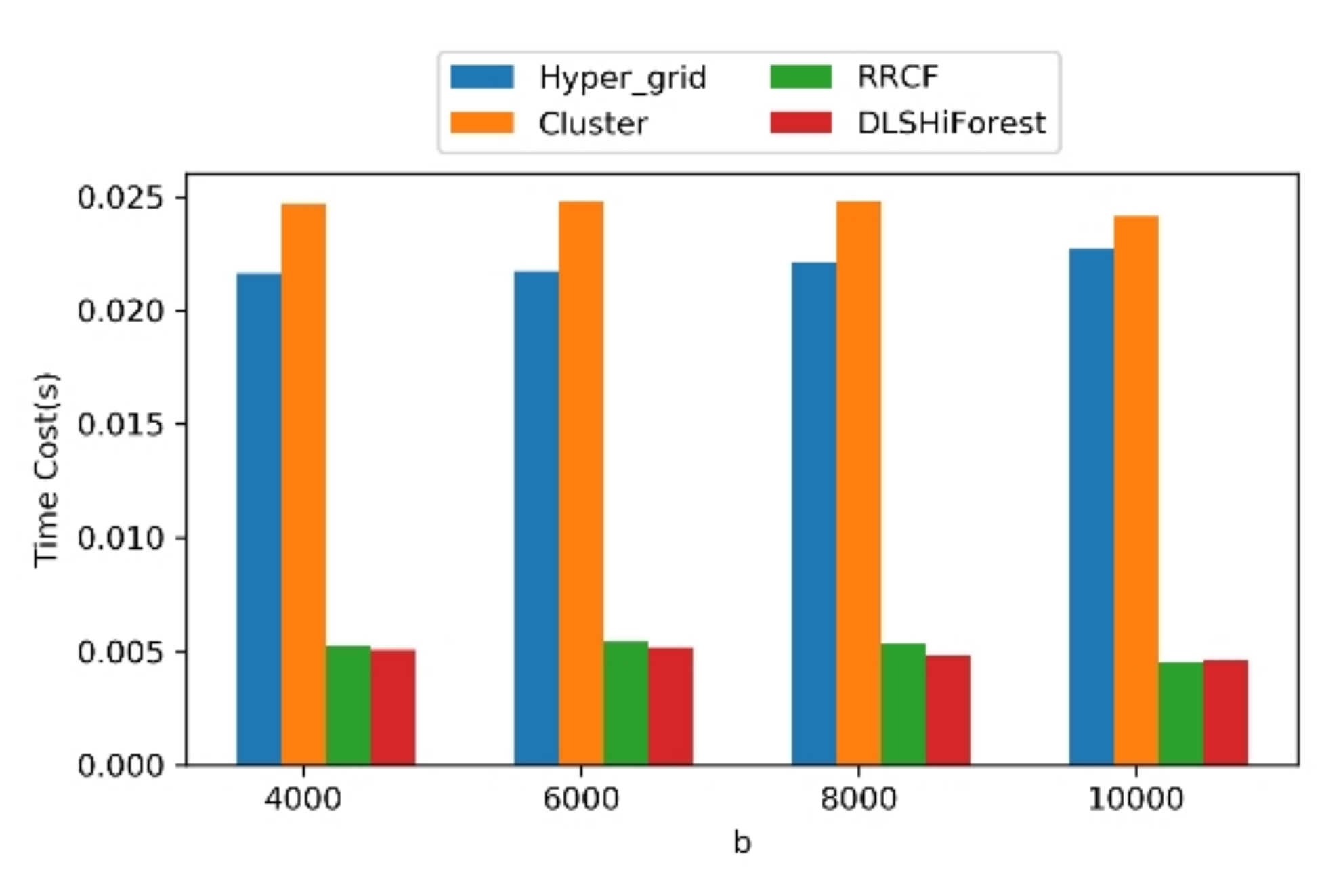}
	\caption*{Fig. 5: Detection efficiency comparison}
\end{figure}

\textbf{Profile 3: Detection accuracy of DLSHiForest w.r.t. (w, t)}

In our method, the window size $w$ and the number of trees $t$ have an important effect on the detection accuracy. Thus, we utilize AUC and F1-Score to examine the accuracy of our method with reference to $w$ and $t$. Here, $w$ = $\left\{64, 128, 256, 512\right\}$, $t$ = $\left\{40, 60, 80, 100\right\}$, and $b$ = 10000. Fig.6 shows the experimental outcomes.

\begin{figure}[!htbp]\setcounter{subfigure}{0}
	\centering
	\subfigure[AUC]{
		\includegraphics[width=1.0\linewidth]{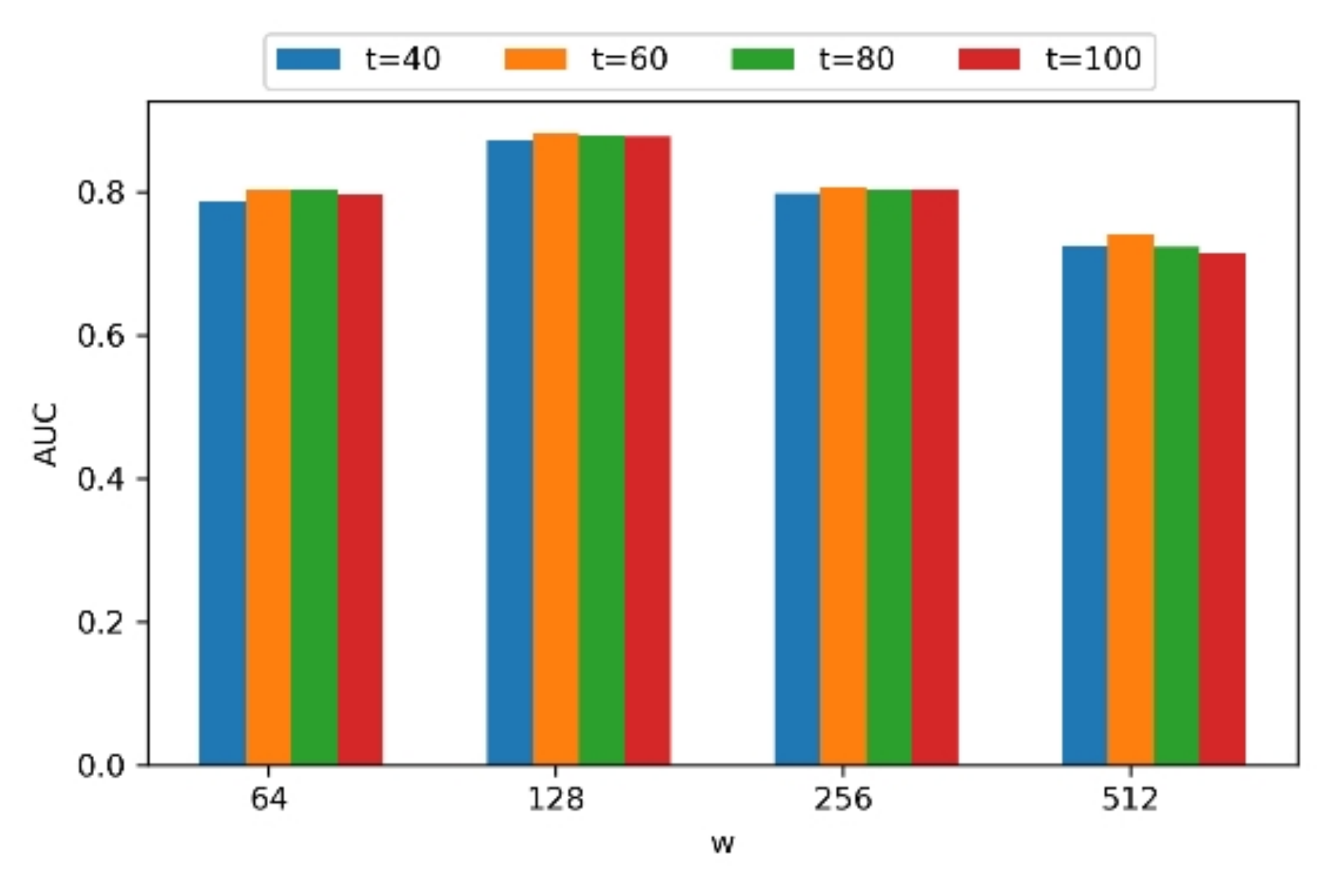}
	}
	\quad
	\subfigure[F1-Score]{
		\includegraphics[width=1.0\linewidth]{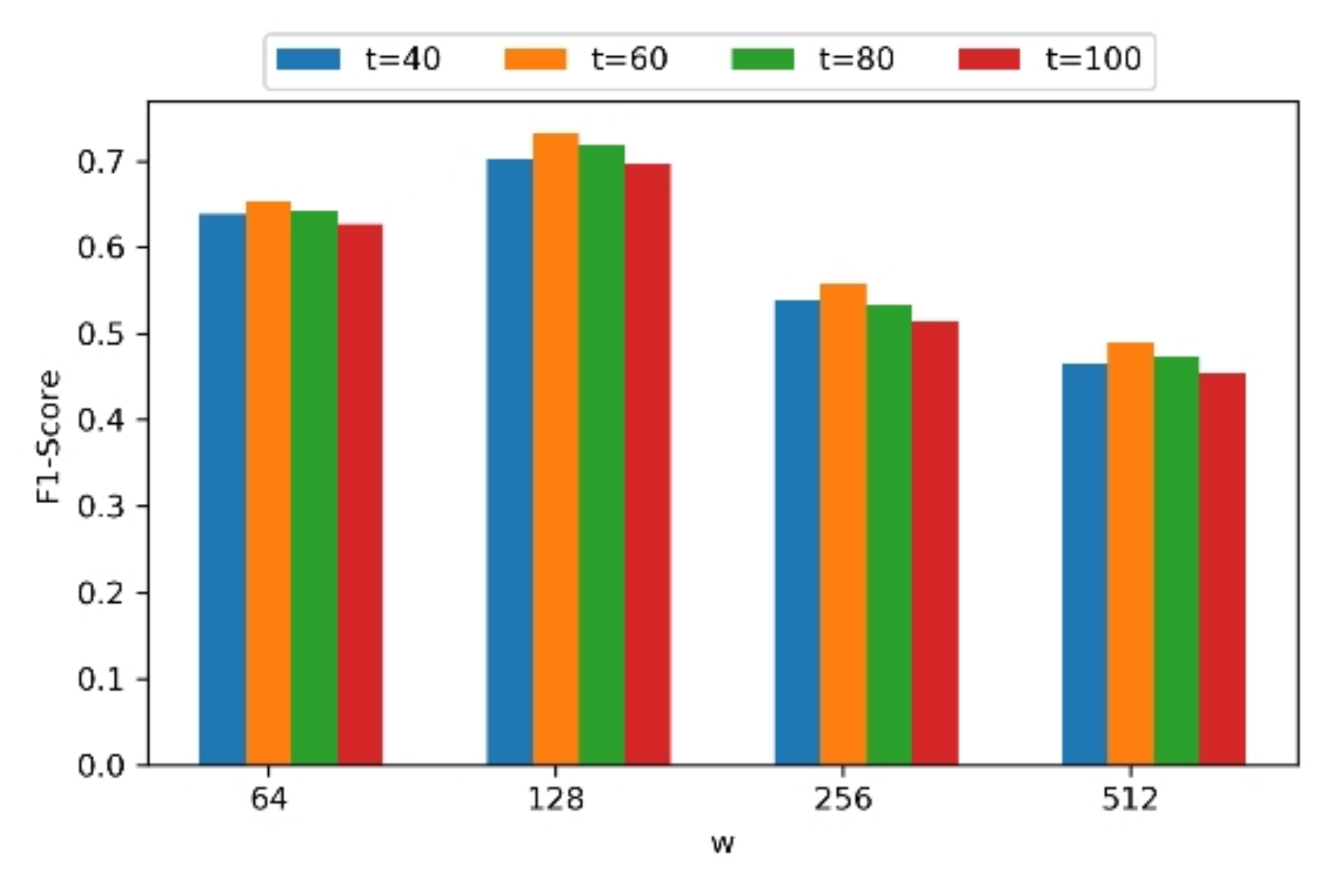}
	}	
	\caption*{Fig. 6: Detection accuracy comparison of DLSHiForest w.r.t. ($w, t$)}
\end{figure}

As shown in Fig.6, when the value of $w$ remains unchanged and $t$ is on the rise, both AUC and F1-Score shows a trend of increasing, then decreasing. The value of number of tree is 60 when the accuracy of our method performs best. This is because the detector consists of multiple weak detectors, i.e., the tree. The performance get better as the $t$ grows. And the detection performance reaches reliably when the number of tree increases to a desired value. While $t$ is constant, AUC and F1-Score also firstly increase, then decrease as the growing of window size. The value of window size is 128 when the performance of our method is better. It is obvious that the window size have an effect on the accuracy of our method. Besides, both AUC and F1-Score reach the maximum while window size and number of trees are 128, 60 respectively, which are used in profile 1 and profile 2 to compare with other methods.

\textbf{Profile 4: Detection efficiency of DLSHiForest w.r.t. (w, t)}

The detection efficiency of proposed method DLSHiForest in respect of $w$ and $t$ is measured. Here, $w$ = $\left\{64, 128, 256, 512\right\}$, $t$ = $\left\{40, 60, 80, 100\right\}$, and $b$ = 10000. Experimental results are shown in Fig.7.

Fig.7 indicates that the efficiency of our method has an upward trend while $w$ is constant and $t$ rises, which is because the data point traverses more trees as $t$ grows. When window size increases and the number of trees remains unchanged, time cost has little fluctuation, which is attributed to the reason that the window size makes a less effect on efficiency. On the whole, our method always maintains a low time cost.

\begin{figure}[!htbp]
	\centering
	\includegraphics[width=1.0\linewidth]{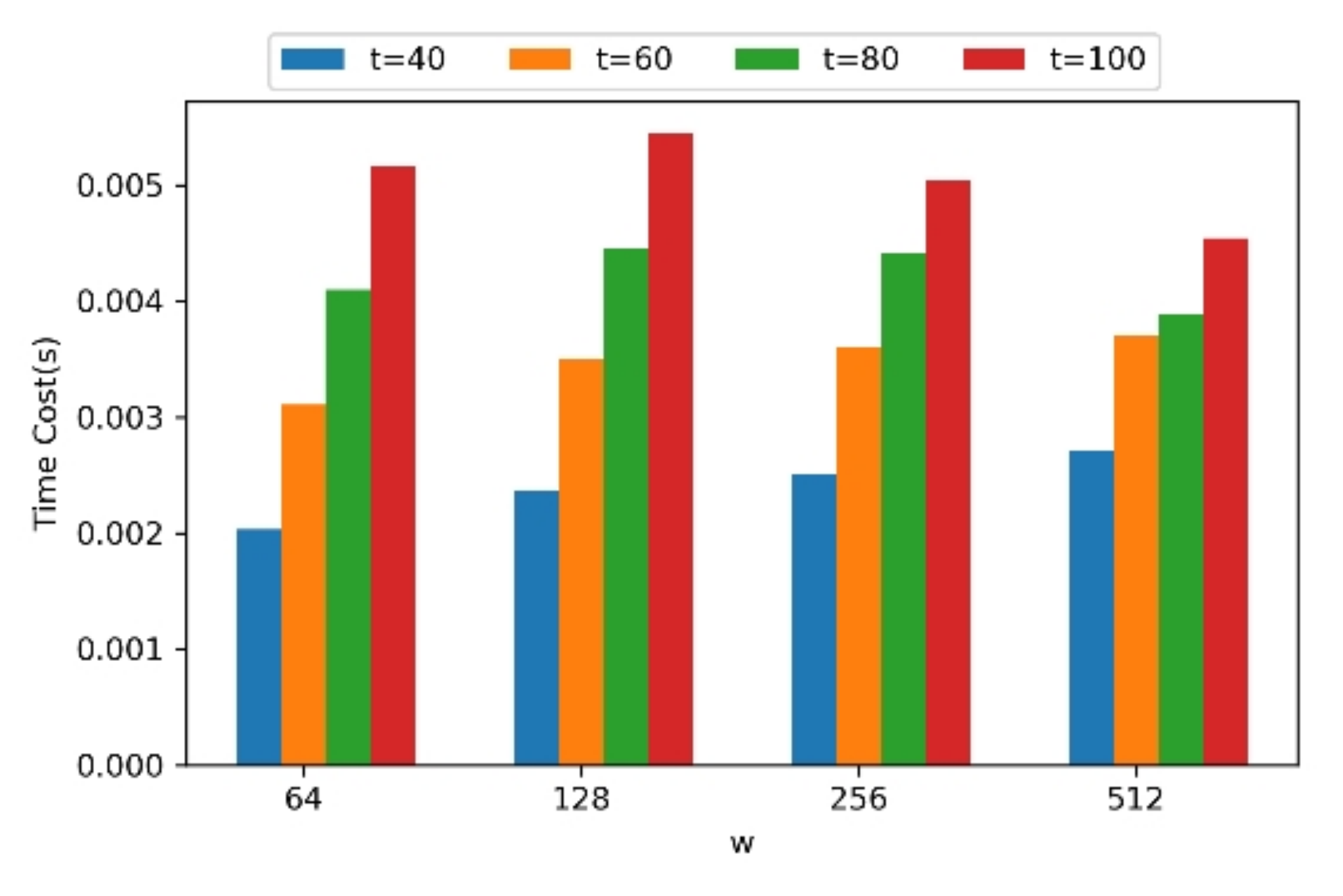}
	\caption*{Fig. 7: Detection efficiency comparison of DLSHiForest w.r.t. ($w, t$)}
\end{figure}

\textbf{Profile 5: The convergence of AUC and F1-Score of DLSHiForest w.r.t. experiment times}

\begin{figure*}[htbp]
	\centering 
	\subfigure{ 
		\includegraphics[width=2.5in]{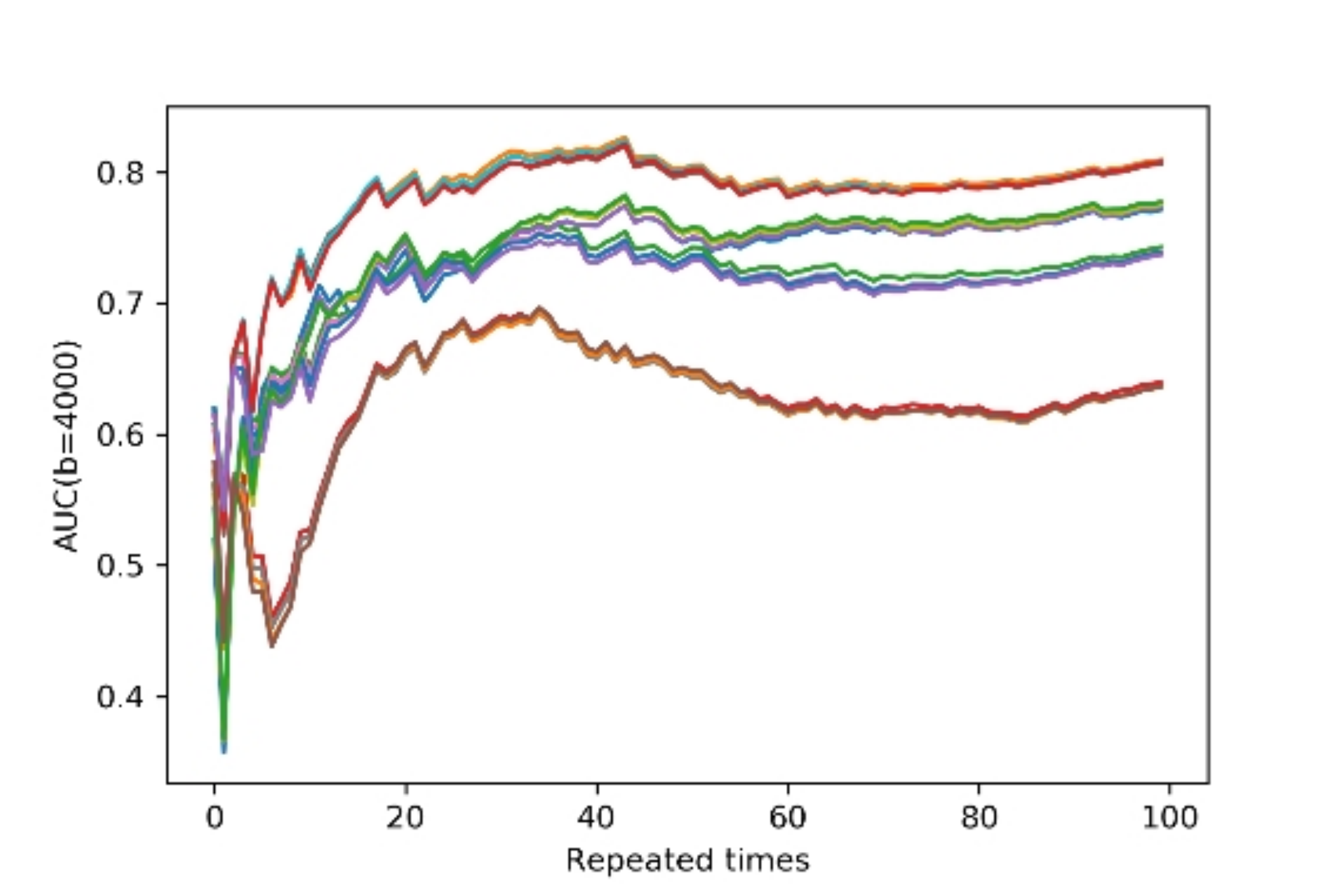} 
	} 
	\subfigure{ 
		\includegraphics[width=2.5in]{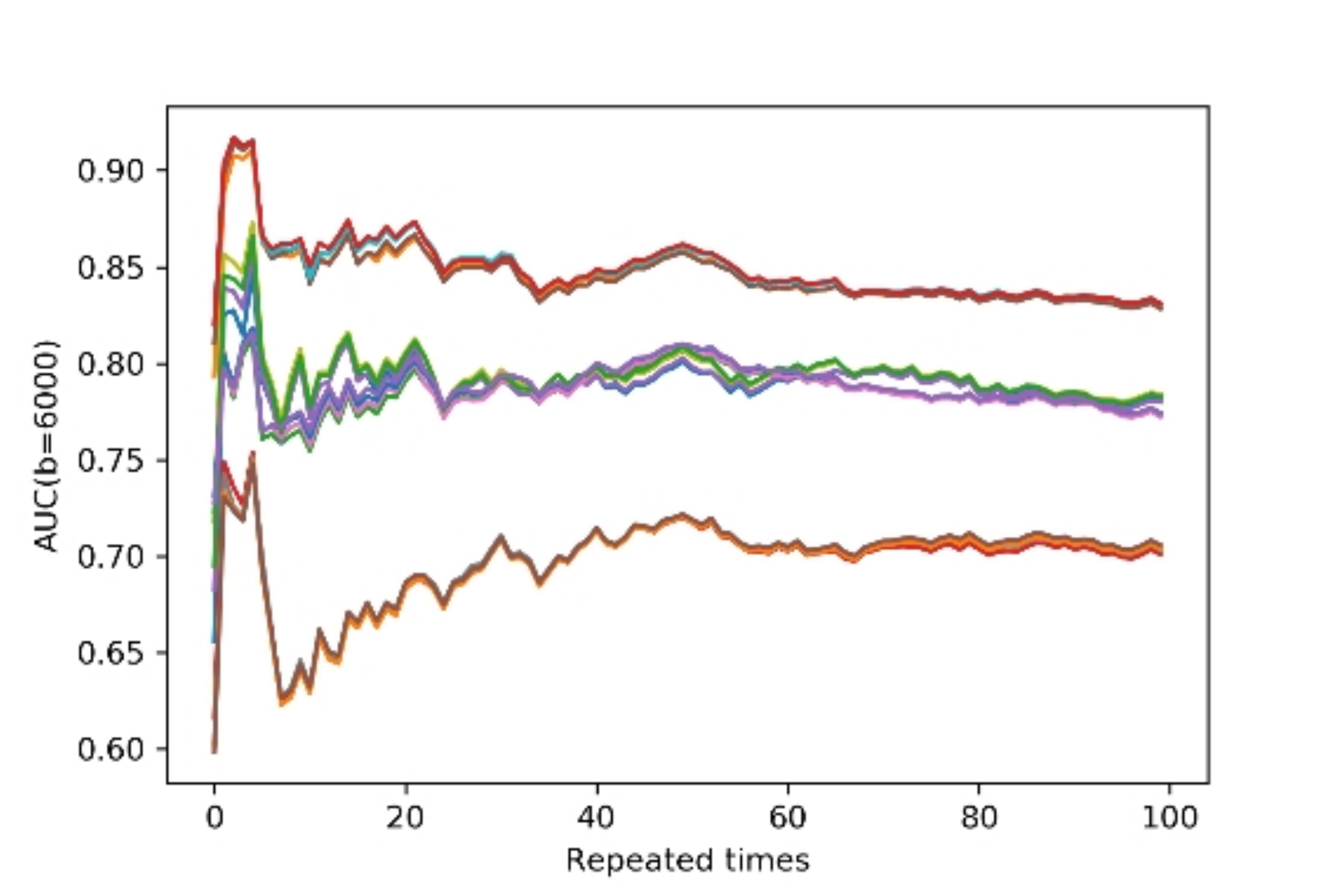} 
	} \\
	\subfigure{ 
		\includegraphics[width=2.5in]{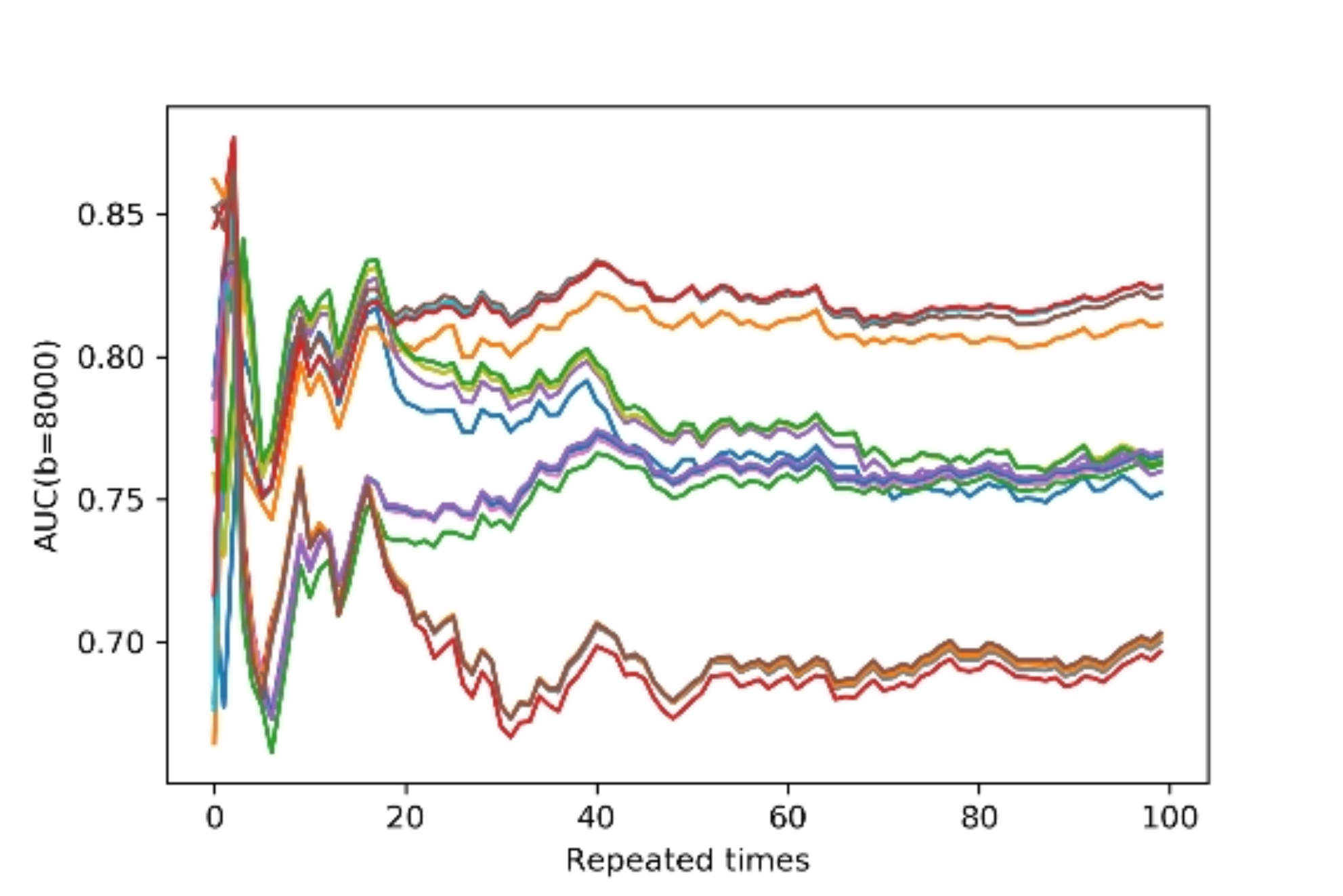} 
	} 
	\subfigure{ 
		\includegraphics[width=2.5in]{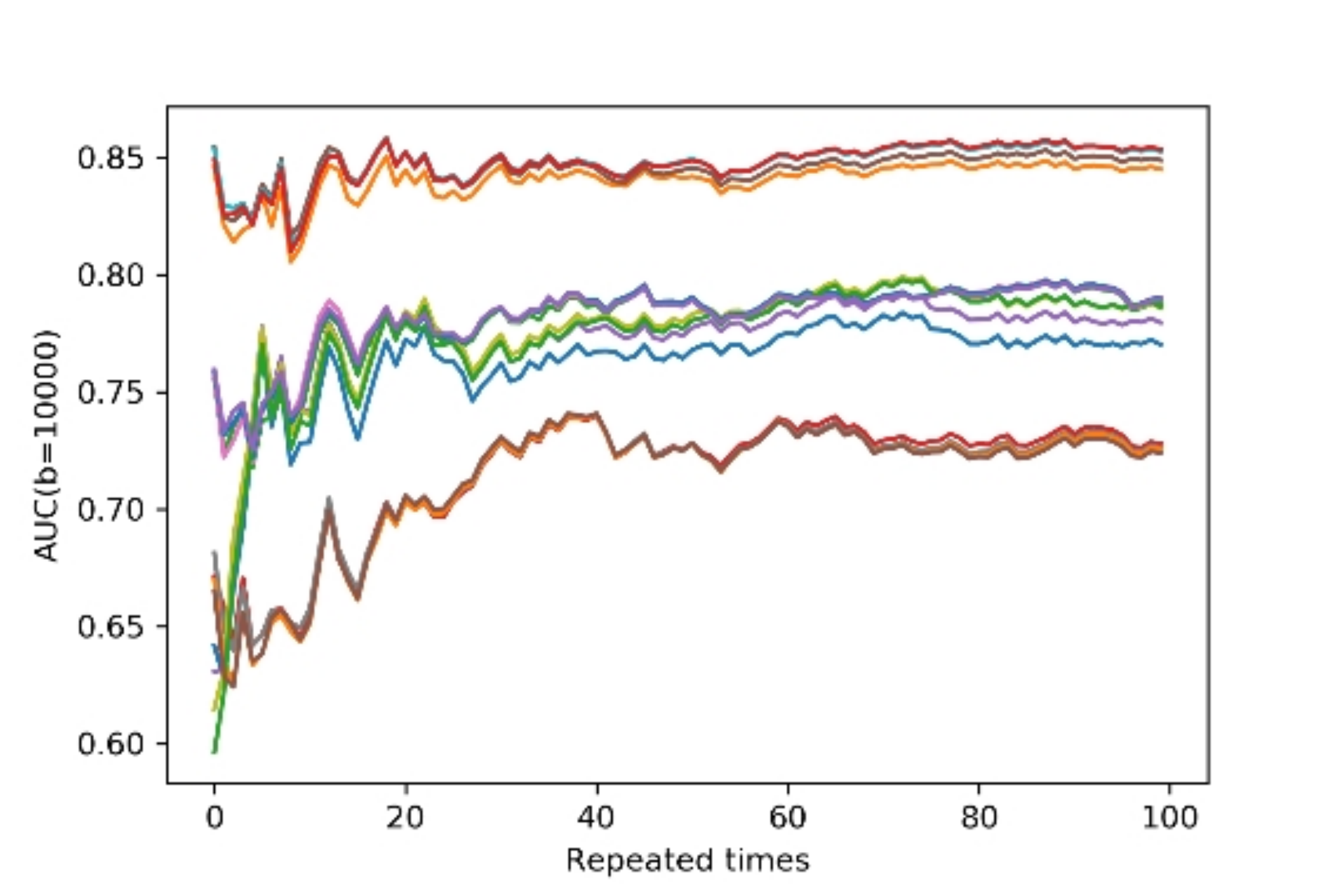} 
	} 
	\caption*{(a) AUC} 
	
	\subfigure{ 
		\includegraphics[width=2.5in]{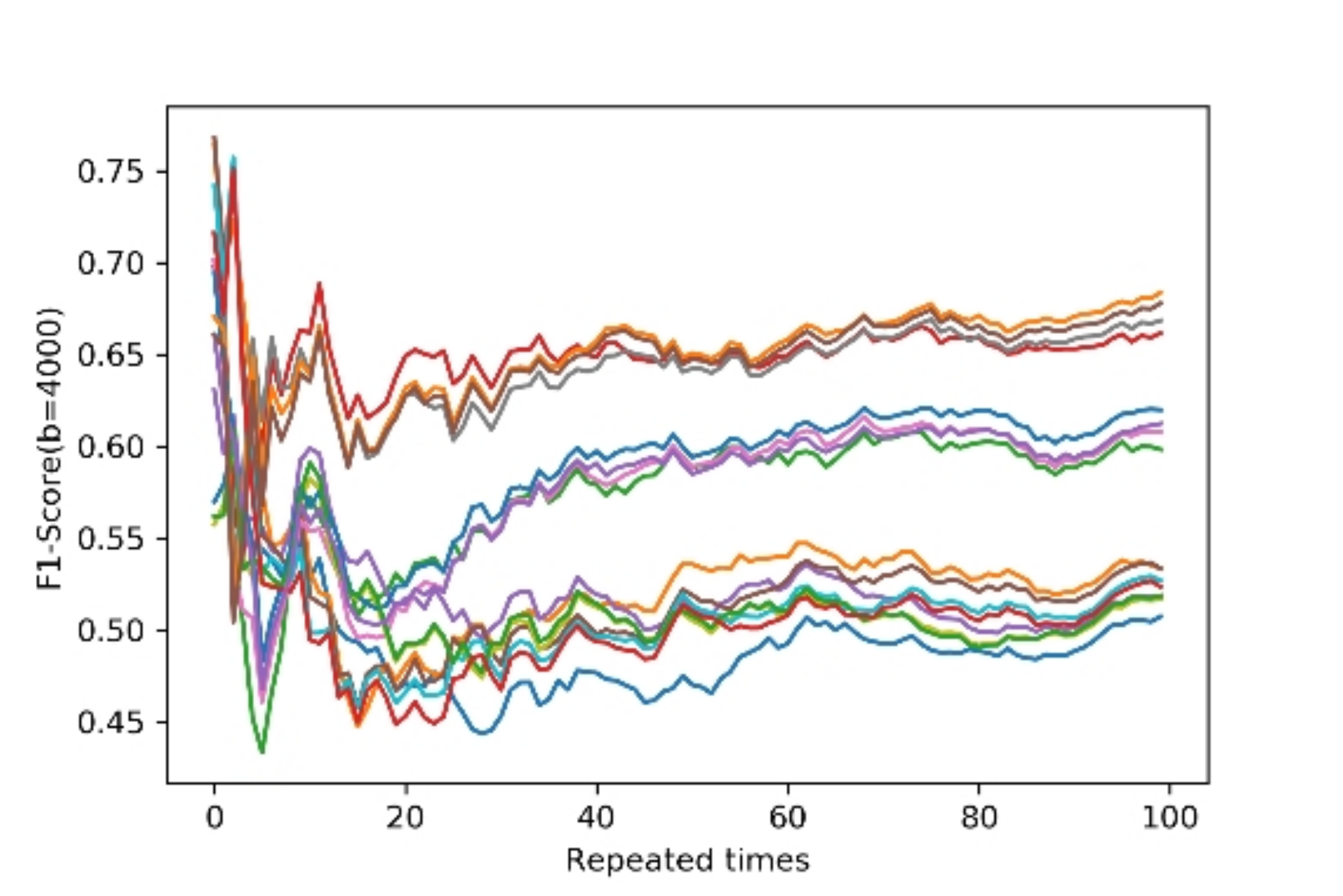} 
	} 
	\subfigure{ 
		\includegraphics[width=2.5in]{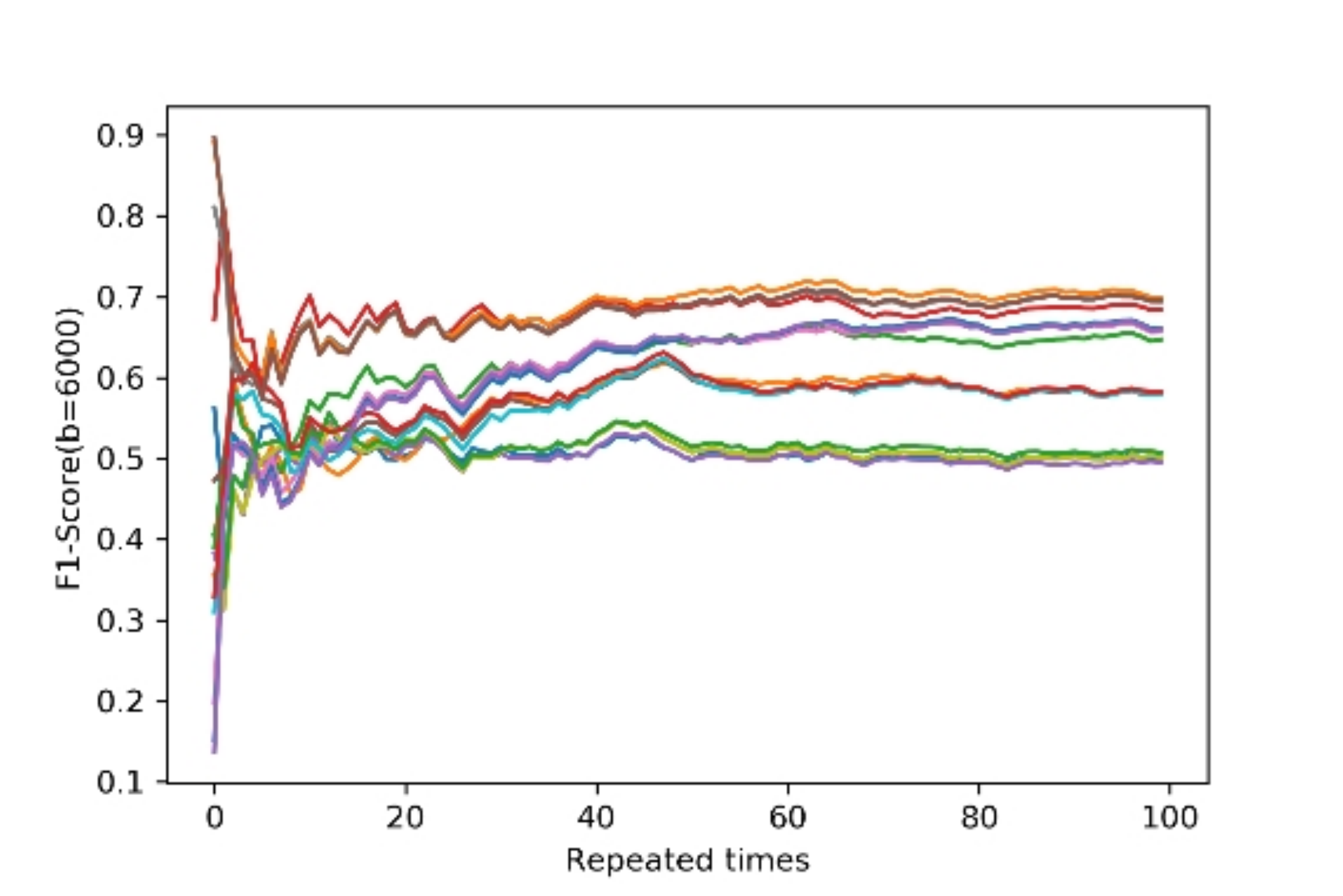} 
	} \\
	\subfigure{ 
		\includegraphics[width=2.5in]{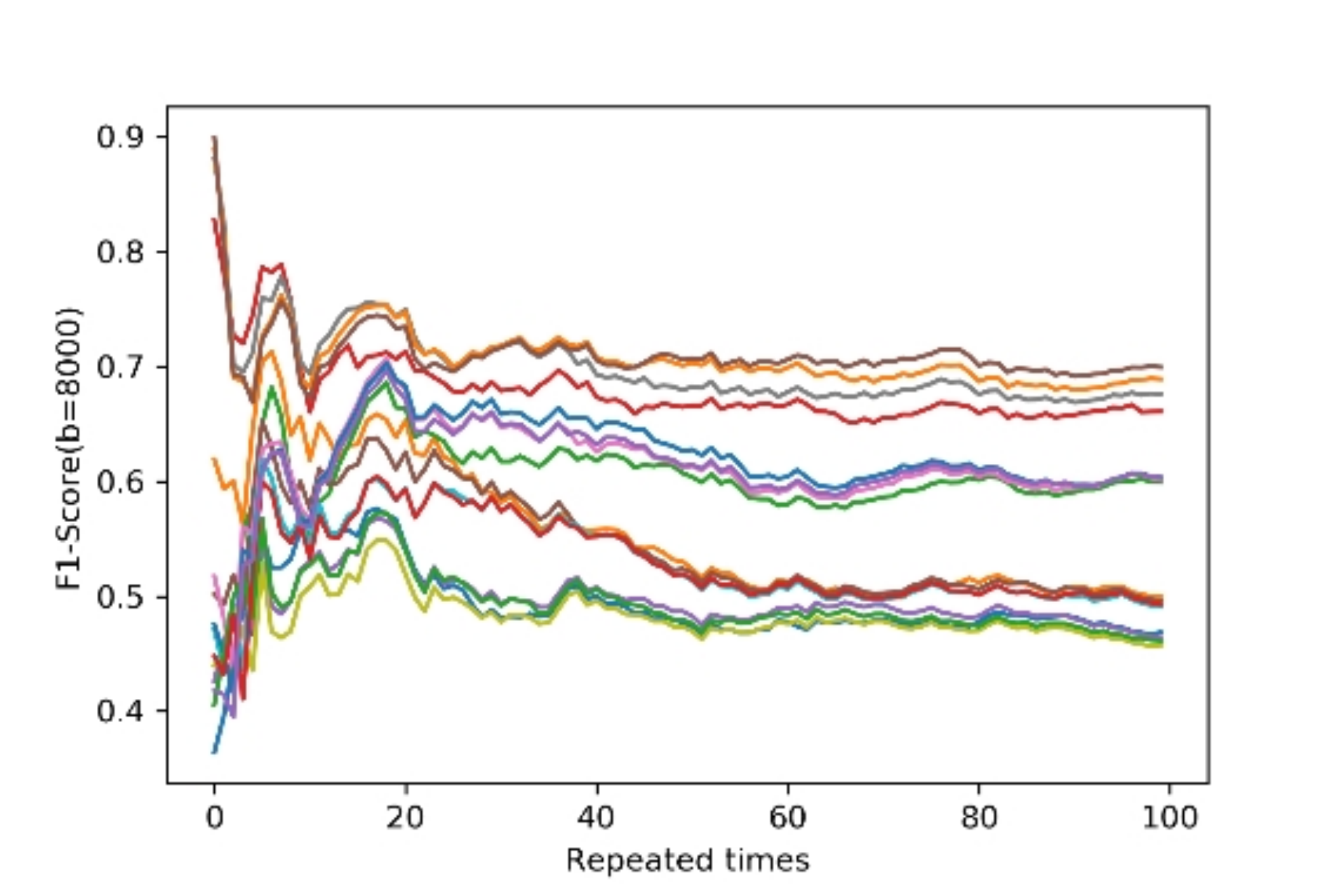} 
	} 
	\subfigure{ 
		\includegraphics[width=2.5in]{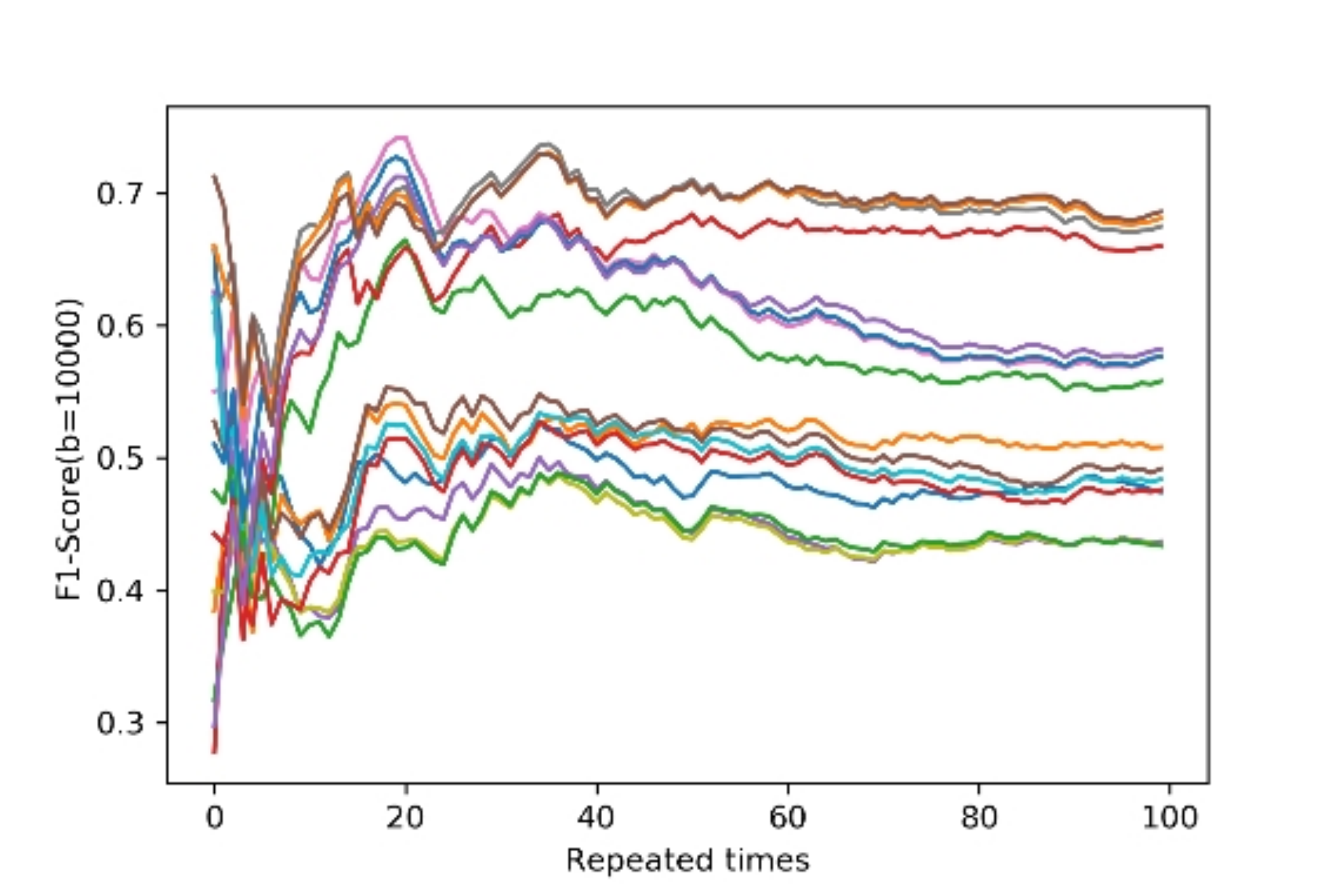} 
	} 
	\caption*{(b) F1-Score} 
	\caption*{Figure 8. AUC/F1-Score convergence of DLSHiForest w.r.t. repeated times.} 
\end{figure*}

The convergence of AUC and F1-Score of our approach DLSHiForest in relation to experiment times is assessed. The experiment times vary from 1 to 100. Fig.8 presents the results. Here, each line denotes an ($w, t$) parameter pair, which includes 16 parameter pairs, and $w$ falls in $\left\{64, 128, 256, 512\right\}$; $t$ belongs to $\left\{40, 60, 80, 100\right\}$; $b$ = $\left\{4000, 6000, 8000, 10000\right\}$.

As Fig.8 shows, DLSHiForest performs well in AUC and F1-Score convergences while experiment times grows. The AUC and F1-Score get relative stable as each experiment is run repeatedly more than 50 times. Therefore, we run each experiment repeatedly 60 times, and average results are utilized eventually. Besides, our method has the better performance when the value of $b$ is 10000, so we set $b$ = 10000 in profile 3 and profile 4 to find the optimal ($w, t$) parameter pair.

\section{Conclusions}
Detecting anomalies in data stream is an important and challenging task. In this paper, current anomaly detection methods have been investigated, and their limitations that they don’t consider the features of infiniteness, correlations and concept drift while tackling streaming data have been indicated, which may lead to low accuracy and efficiency. With the intention of coping with these difficulties, we have put forward the approach DLSHiForest to boost the accuracy and efficiency of detection as taking into account the characteristics of data stream. Ultimately, massive experiments are implemented on real-world agricultural greenhouse dataset. And our experimental results reveal that the accuracy and efficiency of our approach perform well.

In this paper, our model always updates at each time window, which may result in insufficient improvement in aspect of detection performances. In the future, we will consider selective model update scheme, which updates the model while detecting change in data distribution, so as to make better anomaly detection performances.

\section*{Acknowledgements}
This work was supported in part by the Fundamental Research Funds for the Central Universities under Grant No. 30919011282.

%


\bibliographystyle{elsarticle-num}
\bibliography{egbib}
~~~\\
~~~\\







\end{document}